\pdfoutput=1

\RequirePackage[hyphens]{url}
\documentclass[11pt]{article}

\usepackage[final]{acl}

\usepackage{times}
\usepackage{latexsym}

\usepackage[T1]{fontenc}
\usepackage[utf8]{inputenc}
\usepackage{microtype}
\usepackage{inconsolata}
\usepackage{graphicx}
\usepackage{booktabs}
\usepackage{amsmath}
\usepackage{amssymb}

\title{Confidence and Stability of Global and Pairwise Scores in NLP Evaluation}

\author{Georgii Levtsov\thanks{The work was done during the author's internship at JetBrains.} \\
  Neapolis University Pafos / JetBrains \\
  \texttt{g.levtsov.1@nup.ac.cy} \\\And
  Dmitry Ustalov \\
  JetBrains \\
  \texttt{dmitry.ustalov@jetbrains.com} \\}

\begin{document}

\maketitle

\begin{abstract}
With the advent of highly capable instruction-tuned neural language models, benchmarking in natural language processing (NLP) is increasingly shifting towards pairwise comparison leaderboards, such as LMSYS Arena, from traditional global pointwise scores (e.g., GLUE, BIG-bench, SWE-bench). This paper empirically investigates the strengths and weaknesses of both global scores and pairwise comparisons to aid decision-making in selecting appropriate model evaluation strategies. Through computational experiments on synthetic and real-world datasets using standard global metrics and the popular Bradley--Terry model for pairwise comparisons, we found that while global scores provide more reliable overall rankings, they can underestimate strong models with rare, significant errors or low confidence. Conversely, pairwise comparisons are particularly effective for identifying strong contenders among models with lower global scores, especially where quality metrics are hard to define (e.g., text generation), though they require more comparisons to converge if ties are frequent. Our code and data are available at \url{https://github.com/HSPyroblast/srw-ranking} under a permissive license.
\end{abstract}

\section{Introduction}

Modern natural language processing (NLP) benchmarks are often represented as pairwise comparison leaderboards, as seen in projects like LMSYS~Arena \citep{Chiang:24} and AlpacaEval \citep{Dubois:24}. This trend has emerged due to the development of highly capable instruction-tuned large language models (LLMs) that output \emph{textual} rather than categorical responses on open-ended questions. Earlier methods could be reasonably evaluated using static datasets or individual benchmarks. However, modern methods require up-to-date benchmarks that incorporate live feedback from both humans and machines \citep{Faggioli:24}. Previous benchmarks, such as GLUE \citep{Wang:19}, BIG-bench \citep{Srivastava:23}, and SWE-bench \citep{Jimenez:24} or its live-benchmark versions, relied on global pointwise scores, prompting further research into the best approach for NLP benchmarking. But what method is most effective, and in which cases?

In this work, we empirically examine the strengths and weaknesses of pairwise comparisons and global scores. The \emph{goal} of this study is to aid decision-making in selecting the appropriate model evaluation approach, which leads to the two following \emph{research questions}:
\begin{description}
  \item[RQ1.] What are the strengths and limitations of global and pairwise evaluation criteria?
  \item[RQ2.] Which approach is more suitable~for classification problems with binary outputs and for problems where decision values (logits) or textual outputs are available?
\end{description}

To address these research questions, we conducted a series of computational experiments using both synthetic and realistic datasets that were distributed under permissive licenses and included model decision scores. For global evaluation scores, we selected metrics that are widely used in natural language processing and other machine learning tasks. These include accuracy, F-score, and the area under the receiver operating characteristic curve (ROC AUC) for classification tasks, as well as character-level F-score \citep[chrF]{popovic-2015-chrf}, edit distance (ED) \emph{aka} Levenshtein distance, and word error rate (WER) for text generation tasks.

Our findings show that while global scores provide more reliable rankings of models, they tend to underestimate strong models that make rare but significant errors or have modest confidence in their responses. In contrast, pairwise comparisons are particularly effective for identifying strong models among those with relatively low overall scores, especially in cases where the quality metric is difficult to define---such as in text generation, which has been popularized since the release of highly-capable generative models like GPT-3 \citep{Brown:20} and more advanced models.

The remainder of the paper is organized as follows. In Section~\ref{sec:related}, we review the related work. In Section~\ref{sec:problem}, we outline the background of our study and formulate the problem. In Section~\ref{sec:datasets}, we describe the datasets used in our study. In Section~\ref{sec:rq1}, we examine the scoring stability of pairwise comparisons in the case of similar model outputs (RQ1). In Section~\ref{sec:rq2}, we analyze scoring stability in extreme cases of model confidence (RQ2). In Section~\ref{sec:discussion}, we summarize our findings and provide recommendations for using global scores and pairwise comparisons in model selection. Finally, in Section~\ref{sec:conclusion}, we conclude with final remarks and present a flowchart to guide decision-making. Appendices~\ref{app:jigsaw}, \ref{app:sst5}, and \ref{app:ceval} contain supplementary information about the model scores in different settings that we tried in our work.

\section{\label{sec:related}Related Work}

Earlier work by \citet{Furnkranz:03} was focused on using pairwise comparisons (rankings) to train binary classifiers for ranking tasks, while \citet{Broomell:11} explored the use of pairwise model comparisons to identify groups of tasks where each model performs best. \citet{Maystre:17} shown that an optimal ranking of models can be achieved in a linearithmic number of comparisons, inspired by the quicksort algorithm. \citet{Nariya:23} specifically examined the use of pairwise comparisons for small datasets and studied how individual outliers and confounders impact performance estimates.

In contrast to these studies, our work aimed to identify specific scenarios in which pairwise rankings failed or behaved inconsistently, as well as cases in which they provided valuable insights across different task types, namely text classification and text generation.

\section{\label{sec:problem}Problem Formulation}

Suppose we are given a set of models $M$ and an evaluation dataset $X$, where for each element $x_i \in X$, the ground truth labels $G$ and the model predictions $M_i(x_i)$ are known in advance. Our objective is to establish a partial order on $M$. As is common in NLP, this can be done using either global scores or pairwise comparisons. Examples of global scores include widely-used evaluation metrics such as accuracy, ROC AUC, and F-score, while examples of pairwise comparison methods include \citet{Bradley:52}, \citet{Elo:78}, \citet{Newman:23}, and others. We are interested in understanding the reasons behind differences in rankings produced by various methods, so we can effectively leverage the strengths of these metrics.

\paragraph{Global Scores.} For global scores, a function $f(M_i, G) \to \mathbb{R}$, called an \emph{evaluation score}, assigns a numerical score to each model, and the ranking is determined by a permutation $P$ such that
\begin{equation*}
  f(M_{p_1}, G) \geq f(M_{p_2}, G) \geq \dots \geq f(M_{p_m}, G)\text{.}
\end{equation*}

Note that we conducted our experiments on global scores using evaluation measures implemented in scikit-learn~\citep{Pedregosa:11}, edit distance and word error rate from JiWER \citep{Morris:04}, and chrF from sacreBLEU \citep{post-2018-call} libraries for Python.

\paragraph{Pairwise Comparisons.} For pairwise comparisons, a function $f(T) \to P$ derives a ranking from a sequence of pairwise comparisons $(M_i, M_j, w)$, where $w$ indicates whether $M_i$ wins, $M_j$ wins, or the comparison results in a tie. In our case, each test sample $x_t$ provides $\binom{m}{2}$ pairs of models through an auxiliary function
\begin{equation*}
  g(M_i(x_t), M_j(x_t), G(x_t)) \to \{M_i, M_j, 0\}\text{,}
\end{equation*}
and the resulting comparisons are aggregated into the global score, usually indicating the probability of each model winning against the others.

\begin{table*}[t]
\centering
\begin{tabular}{lcrrr}\toprule
\textbf{Dataset} & \textbf{Response} & \textbf{\# of examples} & \textbf{\# of methods} & \textbf{\# of pairs} \\\midrule
Jigsaw \citep{Adams:17} & Categorical & 63,812 & 9 & 2,297,232 \\
SST-5 \citep{socher-etal-2013-recursive} & Categorical & 2,210  & 8 & 61,880 \\
CEval \citep{nguyen-etal-2024-ceval-benchmark} & Textual & 488    & 6 & 7,320 \\\bottomrule
\end{tabular}
\caption{\label{tab:datasets}Descriptive statistics of the datasets used in our study; note that Jigsaw and SST-5 are classification datasets and CEval is a text generation dataset. Numbers of examples and methods are taken from the original test datasets and the corresponding baselines. The number of generated pairs is added by us.}
\end{table*}

For pairwise comparisons, we used the widely known \citet{Bradley:52} ranking model \emph{aka} BT due to its popularity and simplicity. Although other models such as Borda count \citep{deBorda:81}, Elo rating \citep{Elo:78}, TrueSkill \citep{Herbrich:06}, and Rank Centrality \citep{Negahban:17} are also widely used, we chose BT due it its simplicity and popularity. We intentionally did not use Elo or TrueSkill, as their outcomes depend on the order of comparisons,\footnote{\url{https://www.cip.org/blog/llm-judges-are-unreliable}} which is more appropriate for competitive games than for time-insensitive model evaluation. \citet{Bradley:52} is a probabilistic model that estimates a set of latent parameters $p_1, \ldots, p_m$ such that the probability that model $M_i$ outperforms model $M_j$ is given by
\begin{equation*}
  P(M_i \succ M_j) = \frac{p_i}{p_i + p_j}\text{.}
\end{equation*}

We defined $M_i \succ M_j$ to mean that the output of $i$-th model is closer to the correct answer than that of the $j$-th model. We computed the BT scores considering each tie as a half-win and half-lose for both compared items. In our work, we used the implementation of the model from the Evalica library \citep{ustalov-2025-reliable}.

\section{\label{sec:datasets}Datasets}

We conducted experiments on two classification benchmarks, Jigsaw by Google \citep{Adams:17}\footnote{\url{https://jigsaw.google.com/}} and Stanford Sentiment Treebank \citep{socher-etal-2013-recursive} \emph{aka} SST-5, and on one textual benchmark called CEval \citep{nguyen-etal-2024-ceval-benchmark}; see Table~\ref{tab:datasets} for details. We selected these datasets because they provided model outputs for individual examples (including decision-function values), were widely used in the research community, and were available under permissive licenses. We used only test subsets of all datasets. In addition, we ran a series of trials on synthetic and mixed datasets combining both synthetic and real labels.

For each test instance, we compared the outputs of $m$ different models in a pairwise fashion, yielding $\binom{m}{2}$ model pairs. For each pair, we then drew $12m \log(m)$ comparisons at random with replacement,\footnote{We adopted the linearithmic sampling strategy of \citet{Maystre:17} and found through prototyping that a multiplier of 12 gave the best performance.} or else used all available test instances if their count was smaller. Finally, we applied these sampled comparisons to build a Bradley--Terry ranking of the models.

\paragraph{Jigsaw.} We derived a dataset from a popular binary classification dataset for detecting text toxicity called Jigsaw~\citep{Adams:17}. We collected the submission files for nine different models from the leaderboard published by their authors.\footnote{\url{https://www.kaggle.com/competitions/jigsaw-toxic-comment-classification-challenge/code?competitionId=8076&sortBy=scoreDescending&excludeNonAccessedDatasources=true}} Since the authors did not provide ground-truth responses for the test subset of the dataset, we reconstructed them by taking the majority vote from the model-generated responses. These models included the winning method (TTA + PL), DistilBERT, JMTC-20, NB-SVM, XGBoost, XLM-R Conv1D, XLM-R, XLM-RoBERTa Bayesian, and XLM-RoBERTa. Appendix~\ref{app:jigsaw} contains scores exhibited by these models in several variations of this dataset that we created for our experiments.
Although the Jigsaw suite of benchmarks contained other tasks than toxicity detection, e.g., classification bias detection,\footnote{\url{https://www.kaggle.com/competitions/jigsaw-unintended-bias-in-toxicity-classification/code?competitionId=12500&sortBy=scoreDescending&excludeNonAccessedDatasources=true}} we found similar results on them during prototyping. Thus, we decided not to include them in our study.

\paragraph{SST-5.} We used the Stanford Sentiment Treebank dataset \citep[SST-5]{socher-etal-2013-recursive},\footnote{\url{https://nlp.stanford.edu/sentiment/}} a multi-class benchmark for reviews spanning five sentiment categories. To obtain model predictions, we followed the methodology of \citet{Gosgens:21} and re-ran eight open-source baselines.\footnote{\url{https://github.com/prrao87/fine-grained-sentiment}} These baselines included: dictionary-based methods VADER and TextBlob, traditional machine learning methods like logistic regression and support vector machine (SVM), \emph{fast}Text classifier \citep{joulin-etal-2017-bag}, and deep learning classifiers: BERT and ELMo with Flair \citep{akbik-etal-2019-flair} and fine-tuned BERT with Hugging Face \citep{wolf-etal-2020-transformers}. Appendix~\ref{app:sst5} contains the exhibited scores.

\paragraph{CEval.} For a dataset featuring textual outputs evaluated by non-classification metrics, we employed the CEval benchmark for counterfactual text generation \citep{nguyen-etal-2024-ceval-benchmark},\footnote{\url{https://github.com/aix-group/CEval-Counterfactual-Generation-Benchmark}} which measured models' ability to generate text that reversed the emotional tone of the original English input. In this context, we evaluated six models from the original benchmark: Crest, Crowd, GDBA, LLaMA, Llama~2, and MICE. Appendix~\ref{app:ceval} presents the observed scores.

\begin{table}[t]
\centering
\resizebox{\columnwidth}{!}{\begin{tabular}{cccccc}
\toprule
\textbf{Measure} & \textbf{Acc} & \textbf{AUC} & \textbf{BT} & \textbf{F\textsubscript{1}} & \textbf{BT\textsubscript{bin}} \\
\midrule
\textbf{Acc} & $1.00$ & $0.90$ & $-0.23$ & $0.77$ & $0.93$ \\
\textbf{AUC} & $0.90$ & $1.00$ & $0.03$ & $0.87$ & $0.83$ \\
\textbf{BT} & $-0.23$ & $0.03$ & $1.00$ & $0.22$ & $-0.28$ \\
\textbf{F\textsubscript{1}} & $0.77$ & $0.87$ & $0.22$ & $1.00$ & $0.83$ \\
\textbf{BT\textsubscript{bin}} & $0.93$ & $0.83$ & $-0.28$ & $0.83$ & $1.00$ \\
\bottomrule
\end{tabular}}
\caption{\label{tab:jigsaw}\citet{Spearman:04} correlations between model scores in Jigsaw \citep{Adams:17}.}
\end{table}

\section{\label{sec:rq1}Sensitivity to Distributions of Decision Values}


Our first point of interest was focused on the sensitivity of aggregated pairwise comparisons compared to global scores (RQ1). How can we estimate the sensitivity of these evaluations? What occurs when the models exhibit similar performance?

We investigated this by running experiments on the Jigsaw dataset (binary classification) and on SST-5 (multi-class classification). We then examined the decision values of models and used the class with the highest decision value as the model's output.



\paragraph{Raw Decision Values.} We compared the nine Jigsaw models using accuracy (Acc), ROC AUC (AUC), Bradley--Terry (BT) and F\textsubscript{1} scores. For SST-5, we measured F\textsubscript{1}, accuracy and pairwise comparisons, treating the model with the higher confidence score in each pairing as the winner. Table~\ref{tab:jigsaw} showed that the global scores (Acc, AUC, F\textsubscript{1}) yielded consistent, highly correlated rankings, as indicated by the \citet{Spearman:04} correlation coefficient.

On Jigsaw, we found that the anomalous BT ranking resulted from some models, such as XGBoost, outputting only decision values of $0$ or $1$. This caused them to win disproportionately in pairwise comparisons and thus distorted the BT ordering. We observed the same effect on SST-5: SVM rose to the top of the Bradley--Terry ranking due to its more extreme confidence scores, even though its F\textsubscript{1} score lagged behind Flair-BERT, Flair-ELMo, or Transformer. Therefore, \textbf{we recommend applying pairwise comparisons only to models whose decision values share a similar domain}.


\begin{table}[t]
\centering
\begin{tabular}{cccccc}
\toprule
\textbf{Measure} & \textbf{Acc} & \textbf{BT} & \textbf{F\textsubscript{1}}  & \textbf{BT\textsubscript{bin}} \\
\midrule
\textbf{Acc} & $1.00$ & $0.90$ & $0.83$ & $0.69$ \\
\textbf{BT} & $0.90$ & $1.00$ & $0.93$ & $0.55$\\
\textbf{F\textsubscript{1}} & $0.83$ & $0.93$ & $1.00$ & $0.71$\\
\textbf{BT\textsubscript{bin}} & $0.69$ & $0.55$ & $0.71$ &$1.00$ \\
\bottomrule
\end{tabular}
\caption{\label{tab:sst5}\citet{Spearman:04} correlations between model scores in SST-5 \citep{socher-etal-2013-recursive}.}
\end{table}


\paragraph{Binarized Decision Values.} To evaluate our recommendation, we transformed the score-based outputs from Jigsaw and SST-5 into binary values by assigning $1$ to each model's most confident response and $0$ to all others, i.e., by rounding each output to the nearest integer.

This transformation yielded an 88\% fraction of ties on Jigsaw, which affected the rankings derived from pairwise comparisons (denoted as BT\textsubscript{bin} in Table~\ref{tab:jigsaw}), but did not change any of the rankings build using global scores. On SST-5, we observed strong correlations among accuracy, F\textsubscript{1}, and BT rankings (Table~\ref{tab:sst5}), and the ordering remained stable~across different random samples of pairs. Unlike Jigsaw, the larger number of classes on SST-5 resulted in a moderate proportion of ties (about two-thirds of all comparisons), which in turn contributed to the stability of the pairwise rankings. From these experiments, we concluded that \textbf{pairwise comparisons were sensitive to the distributions of decision values across the compared models}.


\paragraph{Binary Responses.} We simulated a binary classification task to examine how binary responses influenced pairwise comparisons and global scores. Three models each produced uniform random binary outputs 1,000 times using different random seeds. An ideal evaluation metric would not have favored any model. We found that accuracy, ROC AUC and F\textsubscript{1} each equaled $0.5$, whereas aggregated \textbf{pairwise comparisons systematically favored one specific model} due to its larger number of evaluated pairs. \citet{Spearman:04} correlation among all global scores was $1$, while the Bradley--Terry ranking exhibited a strong inverse correlation of $-0.5$. These results suggested that pairwise comparison methods were ill-suited for distinguishing between highly similar (or identical) models.

\section{\label{sec:rq2}Instability with Overly Confident Models}

\begin{table}[t]
\centering
\begin{tabular}{ccc}
\toprule
\textbf{Measure} & \textbf{Binary AP} & \textbf{Penalized AP} \\
\midrule
\textbf{MAE} & $0.38$ & $0.86$ \\
\textbf{AUC} & $0.90$ & $0.94$ \\
\textbf{BT} & $[0.33, 0.34]$ & $[0.59, 0.66]$ \\
\textbf{F\textsubscript{1}} & $0.50$ & $0.50$ \\
\bottomrule
\end{tabular}
\caption{\label{tab:confidence}Performance metrics on the adjusted decision functions in the Jigsaw dataset \citep{Adams:17}.}
\end{table}

\begin{table}[t]
\centering
\begin{tabular}{ccc}
\toprule
\textbf{Measure} & \textbf{Binary AP} \\
\midrule
\textbf{ACC} & $1$ \\
\textbf{BT} & $[0.70, 0.71]$ \\
\textbf{F\textsubscript{1}} & $0.5$ \\
\bottomrule
\end{tabular}
\caption{\label{tab:confidence_sst5}Performance metrics on the adjusted decision functions in the SST-5 dataset \citep{socher-etal-2013-recursive}.}
\end{table}

\begin{table}[t]
\centering
\begin{tabular}{ccc}
\toprule
\textbf{Measure} & \textbf{Penalized AP} \\
\midrule
\textbf{ED} & $0.37$ \\
\textbf{WER} & $0.38$ \\
\textbf{chrF} & $0.66$ \\
\textbf{BT} & $[0.66, 0.70]$ \\
\bottomrule
\end{tabular}
\caption{\label{tab:confidence_CEval}Performance metrics on the adjusted decision functions in the CEval dataset \citep{nguyen-etal-2024-ceval-benchmark}.}
\end{table}

Our second point of interest focused on the stability of pairwise comparisons given varying model confidence in the positive class (RQ2). Instead of calculating accuracy, we computed the mean absolute error (MAE) between the binary label of the target class and the model's decision value.



\paragraph{Binarized Decision Values.}  We inflated the confidence of model decision values in the Jigsaw dataset through binarization to assess its impact on model rankings. A good evaluation score should distinguish the original models from the binarized ones, ideally ranking the originals at the top and the binarized models at the bottom.

In the Jigsaw experiments, we observed that under MAE and AUC metrics, most binarized models fell in the rankings according to the average precision score \citep{Buckley:00}. However, based on F\textsubscript{1}, the binarized models received identical scores to the originals due to the binarization performed internally inside the models. In contrast, the Bradley--Terry rankings were disrupted by the inflated model confidences (see Table~\ref{tab:confidence}, Binary AP). Confidence intervals for the Bradley–Terry model, here and throughout the paper, were estimated as 95\% intervals by drawing 1,000 random subsamples of $12m \log(m)$ match sets for each model pair.

Although increased model confidence might challenge the evaluation in text generation tasks, in practice \textbf{it seems difficult to alter textual outputs in a way that changed pairwise rankings without also affecting other evaluation metrics}. In the CEval experiments, both WER and chrF scores remained correlated with the Bradley--Terry pairwise rankings, even after simple manipulations such as appending random strings to the outputs (see Table~\ref{tab:CEval}).

\begin{table}[b]
\centering
\begin{tabular}{cccccc}\toprule
\textbf{Measure} & \textbf{ED} & \textbf{WER} & \textbf{chrF}  & \textbf{BT} \\\midrule
\textbf{ED} & $1.00$ & $0.94$ & $-0.94$ & $-0.94$ \\
\textbf{WER} & $0.94$ & $1.00$ & $-1.00$ & $-0.89$\\
\textbf{chrF} & $-0.94$ & $-1.00$ & $1.00$ & $0.89$\\
\textbf{BT} & $-0.94$ & $-0.89$ & $0.89$ &$1.00$ \\
\bottomrule
\end{tabular}
\caption{\label{tab:CEval}\citet{Spearman:04} correlations between model scores in CEval \citep{nguyen-etal-2024-ceval-benchmark}. Note that some values are negative due to inverted rankings.}
\end{table}





\paragraph{Penalized Decision Values.} In this experiment, we artificially perturbed the model outputs in the Jigsaw and CEval datasets using the ground-truth responses to generate a heavier tail of incorrect answers and to assess how the rankings responded to such perturbations.

For the Jigsaw dataset, we binarized the decision value whenever the model made a mistake, similarly to the previous experiment; otherwise, we left the decision values unchanged. Hence, any mistake led to a model receiving worse scores, while models without errors retained their original scores. We found that under MAE and AUC, most penalized models fell to the bottom of the rankings, whereas F\textsubscript{1} produced results identical to those of the earlier experiment. The Bradley--Terry rankings did not correlate well with the other metrics; nevertheless, they correctly placed most original models above the penalized ones (see Table~\ref{tab:confidence}, Penalized AP, and a similar Table~\ref{tab:confidence_sst5} for SST-5).

A similar pattern arose in the text-generation tasks. We appended random long strings to a random 5\% of model outputs in the CEval dataset, which caused their distance-based global scores (ED and WER) to decline, positioning them near the bottom. However, the pairwise and chrF rankings remained largely stable (see Table~\ref{tab:confidence_CEval}, Penalized AP). Given that a 5\% error rate can represent a substantial difference, we recommend filtering out such extreme cases or employing multiple evaluation metrics, since pairwise comparisons tend to be relatively insensitive to rare but large deviations.

From this experiment, we concluded that \textbf{pairwise comparisons can still favor promising models even when they commit rare but significant errors}.

\paragraph{Scored Responses.} As suggested by \citet{Gosgens:21} and confirmed by our experiments, the F\textsubscript{1} score was a viable alternative to accuracy for binary classification tasks with an available decision function. However, ROC AUC and BT yielded more accurate results and recovered the true ranking. Nonetheless, \textbf{pairwise comparisons had to be conducted carefully to avoid favoring models that produced more confident predictions}, e.g., decision values closer to the extremes, like logits near $0$ or $1$.

\begin{figure}[t]
\centering
\includegraphics[width=\columnwidth]{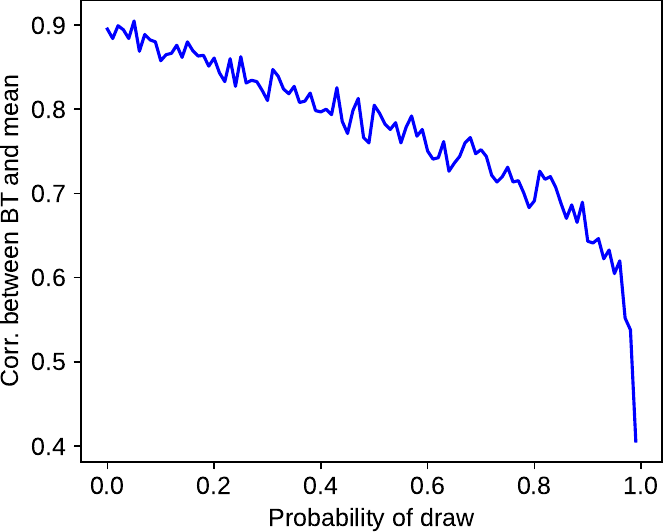}
\caption{\label{fig:draws}Dependency of the correlation between absolute and pairwise rankings in a synthetic experiment based on the CEval dataset \citep{nguyen-etal-2024-ceval-benchmark}. The results show that the Bradley--Terry model produces reliable rankings even with a large fraction of ties.}
\end{figure}

\section{\label{sec:discussion}Discussion}

\paragraph{Draws in Comparisons.} We noticed that \citet{Bradley:52} rankings had performed poorly when a large fraction of comparisons resulted in draws (Section~\ref{sec:rq1}). They produced indistinguishable results and required a high number of observations to achieve a stable~ranking, which led to high computational costs. Accuracy also tended to penalize models that made rare but significant errors. In contrast, pairwise comparisons identified such models effectively, although they sometimes demanded additional measures to ensure correctness (Section~\ref{sec:rq2}). Pairwise comparisons proved particularly useful for tasks which are uneasy to evaluate according to the ground-truth data, as had been confirmed by modern benchmarks \citep{Chiang:24,Dubois:24}.

In text generation tasks, ties occurred far less frequently than in classification, since evaluation metrics for generation rarely yielded identical scores. Using the CEval dataset as an example, we simulated the effect of introducing synthetic ties on the resulting rankings. More specifically, we measured the correlation between average rankings and pairwise chrF-based rankings for five models, varying the tie probability from $0$ to $1$ in increments of $0.01$. For each probability level, we conducted 1,000 trials with $12m \log(m)$ matches per model pair. The results demonstrated that the rankings maintained a strong correlation ($0.8$) even when ties represented up to 50\% of outcomes (see Figure~\ref{fig:draws}).

However, we observed that this behavior generally depended on both the closeness of model performance and the total number of comparisons done.

\begin{figure}[t]
\centering
\includegraphics[width=\columnwidth]{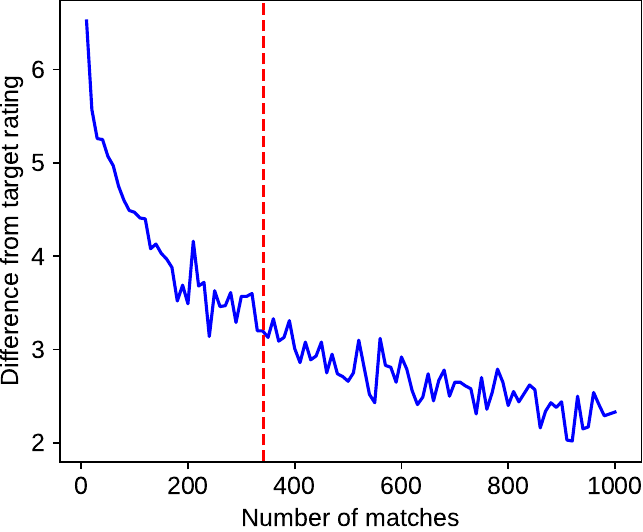}
\caption{\label{fig:stability}Comparison of stability in the Jigsaw dataset \citep{Adams:17}. The red line indicates $12m \log(m)$.}
\end{figure}


\paragraph{Comparison Stability.} To examine how the number of comparisons affects ranking stability, we constructed Bradley–Terry rankings by randomly selecting an equal number of comparisons for each pair of models, varying this number from 10 to 1000 in increments of 10. At each step, we computed the average number of changes in the ranking over 100 trials, relative to the ranking obtained using 100,000 random comparisons per pair. As mentioned earlier, we adopted the linearithmic sampling strategy proposed by \citet{Maystre:17} and settled on using $12m \log(m)$ comparisons, which provided stable results while maintaining a low computational complexity. Figure~\ref{fig:stability} presents the corresponding plot for the Jigsaw dataset, though a similar effect was observed across the other datasets as well.

\begin{figure}[t]
\includegraphics[width=\columnwidth]{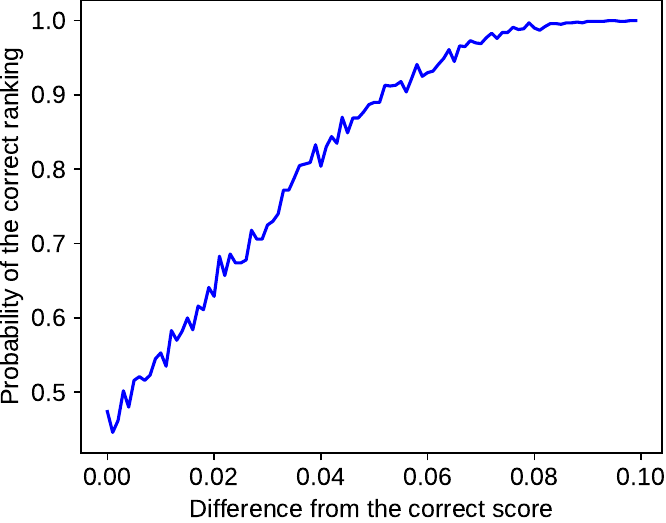}
\caption{\label{fig:differences}Dependency of probability on difference in a synthetic experiment: the larger the difference between model outputs, the better pairwise comparisons can correctly rank the models.}
\end{figure}

\paragraph{Magnitude of Difference.} As in the binary-response experiment described earlier, we investigated the magnitude of differences that aggregated pairwise comparisons could detect. Specifically, we examined how the probability of correct ranking depended on the difference between the decision functions of the models, such as logits or class scores. We created a grid of score differences spanning $0.9$ to $1.0$ in $100$ steps. At each step, we subtracted the value from a randomly selected pair's scores and repeated this procedure 1,000 times. As shown in Figure~\ref{fig:differences}, \textbf{pairwise comparisons perform best when the difference between model outputs is non-negligible}; for example, when there was at least a 10\% difference in class probability in our synthetic example.

\begin{figure*}[t]
\centering
\includegraphics[width=.85\textwidth]{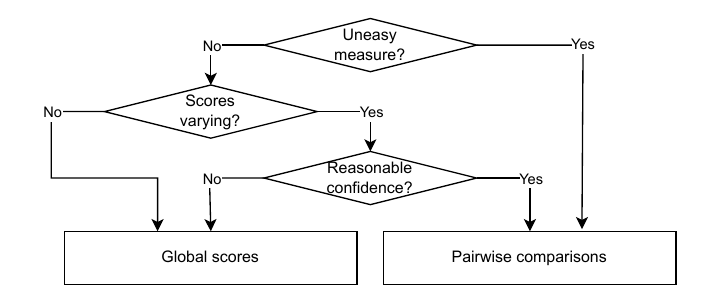}
\caption{\label{fig:decision}How to choose between global scores and pairwise comparisons? Pairwise comparisons are especially effective when the evaluation involves a difficult-to-define (``uneasy'') measure, such as in text generation, or when model scores vary widely and no model shows strong confidence. In contrast, if the measure is clearly defined, the scores are relatively consistent, or some models produce more confident predictions, global evaluation metrics may be a better choice.}
\end{figure*}

\section{\label{sec:conclusion}Conclusion}

Our studies showed that pairwise comparisons identified potentially good models among those with poor global scores. They performed well on problems where the quality measure was difficult to define, such as text generation (RQ2). However, when a large fraction of comparisons ended in ties, the algorithm required a large number of comparisons to converge. In contrast, global scores performed better on evaluation measures that were easier to define and generally required smaller amounts of data (RQ1). Nevertheless, global scores tended to underestimate models that committed rare but significant errors. These results were consistent across synthetic datasets, multiple public datasets, and their variations.

While our study was limited to experiments on only three datasets, we believe the actionable recommendations we have discovered will advance the state of benchmarking in NLP. In addition to replicating our experiments on other datasets with different sets of models, we also find it interesting to explore which subset of the data each model performs best on, where we expect pairwise comparisons to excel. Figure~\ref{fig:decision} presents the flowchart for the model evaluation approach selection. Another possible limitation of our study was the use of well-known NLP datasets released before the wide adoption of LLMs. However, we believe that our results would generalize to newer datasets and models, as we observed the same effects consistently across all datasets, including the relatively recent textual dataset CEval. This analysis included then state-of-the-art open LLMs, such as Llama 2 and LLaMA. Running our experiments on a new multi-task dataset with frontier LLM responses would allow for a more comprehensive evaluation of the observed effects in a modern setting.

Although our experiments had been limited to three datasets, we believe that the actionable recommendations we derived could advance the state of NLP benchmarking. For future work, it would have been useful to replicate our experiments on additional datasets with diverse model sets and to examine the specific data subsets on which each model performed best, anticipating that pairwise comparisons would have excelled in those scenarios.

\section*{Acknowledgments}

The authors are grateful to three anonymous reviewers whose comments allowed us to improve the manuscript. We are also grateful to the anonymous mentor who provided vital feedback during the pre-submission mentorship program at the ACL Student Research Workshop. Last but not least, we are grateful to the Internships and Academy teams at JetBrains for supporting Georgii's work.

\bibliography{anthology,custom}

\begin{thebibliography}{31}
\providecommand{\natexlab}[1]{#1}

\bibitem[{Adams et~al.(2017)Adams, Sorensen, Elliott, Dixon, McDonald, Thain, and Cukierski}]{Adams:17}
CJ~Adams, Jeffrey Sorensen, Julia Elliott, Lucas Dixon, Mark McDonald, Nithum Thain, and Will Cukierski. 2017.
\newblock {Toxic Comment Classification Challenge}.
\newblock \url{https://kaggle.com/competitions/jigsaw-toxic-comment-classification-challenge}.
\newblock Kaggle.

\bibitem[{Akbik et~al.(2019)Akbik, Bergmann, Blythe, Rasul, Schweter, and Vollgraf}]{akbik-etal-2019-flair}
Alan Akbik, Tanja Bergmann, Duncan Blythe, Kashif Rasul, Stefan Schweter, and Roland Vollgraf. 2019.
\newblock \href {https://doi.org/10.18653/v1/N19-4010} {{FLAIR}: An easy-to-use framework for state-of-the-art {NLP}}.
\newblock In \emph{Proceedings of the 2019 Conference of the North {A}merican Chapter of the Association for Computational Linguistics (Demonstrations)}, pages 54--59, Minneapolis, Minnesota. Association for Computational Linguistics.

\bibitem[{Bradley and Terry(1952)}]{Bradley:52}
Ralph~Allan Bradley and Milton~E. Terry. 1952.
\newblock \href {https://doi.org/10.2307/2334029} {{Rank Analysis of Incomplete Block Designs: I. The Method of Paired Comparisons}}.
\newblock \emph{Biometrika}, 39(3/4):324--345.

\bibitem[{Broomell et~al.(2011)Broomell, Budescu, and Por}]{Broomell:11}
Stephen~B. Broomell, David~V. Budescu, and Han-Hui Por. 2011.
\newblock \href {https://doi.org/10.1017/S1930297500004241} {Pair-wise comparisons of multiple models}.
\newblock \emph{Judgment and Decision Making}, 6(8):821--831.

\bibitem[{Brown et~al.(2020)Brown, Mann, Ryder, Subbiah, Kaplan, Dhariwal, Neelakantan, Shyam, Sastry, Askell, Agarwal, Herbert-Voss, Krueger, Henighan, Child, Ramesh, Ziegler, Wu, Winter, Hesse, Chen, Sigler, Litwin, Gray, Chess, Clark, Berner, McCandlish, Radford, Sutskever, and Amodei}]{Brown:20}
Tom~B. Brown, Benjamin Mann, Nick Ryder, Melanie Subbiah, Jared Kaplan, Prafulla Dhariwal, Arvind Neelakantan, Pranav Shyam, Girish Sastry, Amanda Askell, Sandhini Agarwal, Ariel Herbert-Voss, Gretchen Krueger, Tom Henighan, Rewon Child, Aditya Ramesh, Daniel~M. Ziegler, Jeffrey Wu, Clemens Winter, and 12 others. 2020.
\newblock \href {https://proceedings.neurips.cc/paper_files/paper/2020/file/1457c0d6bfcb4967418bfb8ac142f64a-Paper.pdf} {{Language Models are Few-Shot Learners}}.
\newblock In \emph{Advances in Neural Information Processing Systems 33}, NeurIPS~2020, pages 1877--1901, Montr\'{e}al, QC, Canada. Curran Associates, Inc.

\bibitem[{Buckley and Voorhees(2000)}]{Buckley:00}
Chris Buckley and Ellen~M. Voorhees. 2000.
\newblock \href {https://doi.org/10.1145/345508.345543} {{Evaluating Evaluation Measure Stability}}.
\newblock In \emph{Proceedings of the 23rd Annual International ACM SIGIR Conference on Research and Development in Information Retrieval}, SIGIR '00, pages 33--40, Athens, Greece. Association for Computing Machinery.

\bibitem[{Chiang et~al.(2024)Chiang, Zheng, Sheng, Angelopoulos, Li, Li, Zhu, Zhang, Jordan, Gonzalez, and Stoica}]{Chiang:24}
Wei-Lin Chiang, Lianmin Zheng, Ying Sheng, Anastasios~Nikolas Angelopoulos, Tianle Li, Dacheng Li, Banghua Zhu, Hao Zhang, Michael Jordan, Joseph~E. Gonzalez, and Ion Stoica. 2024.
\newblock \href {https://proceedings.mlr.press/v235/chiang24b.html} {{Chatbot Arena: An Open Platform for Evaluating LLMs by Human Preference}}.
\newblock In \emph{Proceedings of the 41st International Conference on Machine Learning}, volume 235 of \emph{Proceedings of Machine Learning Research}, pages 8359--8388. PMLR.

\bibitem[{de~Borda(1781)}]{deBorda:81}
Jean-Charles de~Borda. 1781.
\newblock {M\'{e}moire sur les élections au scrutin}.
\newblock \emph{Histoire de l’Académie royale des sciences}, pages 657--665.

\bibitem[{Dubois et~al.(2024)Dubois, Liang, and Hashimoto}]{Dubois:24}
Yann Dubois, Percy Liang, and Tatsunori Hashimoto. 2024.
\newblock \href {https://openreview.net/forum?id=CybBmzWBX0} {{Length-Controlled AlpacaEval: A Simple Debiasing of Automatic Evaluators}}.
\newblock In \emph{First Conference on Language Modeling}.

\bibitem[{Elo(1978)}]{Elo:78}
Arpad~E. Elo. 1978.
\newblock \emph{{The Rating Of Chess Players, Past \& Present}}.
\newblock Arco Publishing Inc., New York.

\bibitem[{Faggioli et~al.(2024)Faggioli, Dietz, Clarke, Demartini, Hagen, Hauff, Kando, Kanoulas, Potthast, Stein, and Wachsmuth}]{Faggioli:24}
Guglielmo Faggioli, Laura Dietz, Charles L.~A. Clarke, Gianluca Demartini, Matthias Hagen, Claudia Hauff, Noriko Kando, Evangelos Kanoulas, Martin Potthast, Benno Stein, and Henning Wachsmuth. 2024.
\newblock \href {https://doi.org/10.1145/3624730} {{Who Determines What Is Relevant? Humans or AI? Why Not Both?}}
\newblock \emph{Communications of the ACM}, 67(4):31--34.

\bibitem[{F\"{u}rnkranz and H\"{u}llermeier(2003)}]{Furnkranz:03}
Johannes F\"{u}rnkranz and Eyke H\"{u}llermeier. 2003.
\newblock \href {https://doi.org/10.1007/978-3-540-39857-8_15} {{Pairwise Preference Learning and Ranking}}.
\newblock In \emph{Machine Learning: ECML 2003}, volume 2837 of \emph{Lecture Notes in Computer Science}, pages 145--156. Springer.

\bibitem[{G\"{o}sgens et~al.(2021)G\"{o}sgens, Zhiyanov, Tikhonov, and Prokhorenkova}]{Gosgens:21}
Martijn G\"{o}sgens, Anton Zhiyanov, Aleksey Tikhonov, and Liudmila Prokhorenkova. 2021.
\newblock \href {https://proceedings.neurips.cc/paper/2021/file/8e489b4966fe8f703b5be647f1cbae63-Paper.pdf} {{Good Classification Measures and How to Find Them}}.
\newblock In \emph{Advances in Neural Information Processing Systems 34}, NeurIPS~2021, pages 17136--17147, Online. Curran Associates, Inc.

\bibitem[{Herbrich et~al.(2006)Herbrich, Minka, and Graepel}]{Herbrich:06}
Ralf Herbrich, Tom Minka, and Thore Graepel. 2006.
\newblock \href {https://doi.org/10.7551/mitpress/7503.003.0076} {{TrueSkill\textsuperscript{\texttrademark}: A Bayesian Skill Rating System}}.
\newblock In \emph{Advances in Neural Information Processing Systems~19}, pages 569--576, Vancouver, BC, Canada. MIT Press.

\bibitem[{Jimenez et~al.(2024)Jimenez, Yang, Wettig, Yao, Pei, Press, and Narasimhan}]{Jimenez:24}
Carlos~E. Jimenez, John Yang, Alexander Wettig, Shunyu Yao, Kexin Pei, Ofir Press, and Karthik~R. Narasimhan. 2024.
\newblock \href {https://openreview.net/forum?id=VTF8yNQM66} {{SWE-bench: Can Language Models Resolve Real-World GitHub Issues?}}
\newblock In \emph{Proceedings of the Twelfth International Conference on Learning Representations (ICLR)}.

\bibitem[{Joulin et~al.(2017)Joulin, Grave, Bojanowski, and Mikolov}]{joulin-etal-2017-bag}
Armand Joulin, Edouard Grave, Piotr Bojanowski, and Tomas Mikolov. 2017.
\newblock \href {https://aclanthology.org/E17-2068/} {Bag of tricks for efficient text classification}.
\newblock In \emph{Proceedings of the 15th Conference of the {E}uropean Chapter of the Association for Computational Linguistics: Volume 2, Short Papers}, pages 427--431, Valencia, Spain. Association for Computational Linguistics.

\bibitem[{Maystre and Grossglauser(2017)}]{Maystre:17}
Lucas Maystre and Matthias Grossglauser. 2017.
\newblock \href {https://proceedings.mlr.press/v70/maystre17a.html} {{Just Sort It! A Simple and Effective Approach to Active Preference Learning}}.
\newblock In \emph{Proceedings of the 34th International Conference on Machine Learning}, volume~70 of \emph{ICML~2017}, pages 2344--2353, Sydney, NSW, Australia. PMLR.

\bibitem[{Morris et~al.(2004)Morris, Maier, and Green}]{Morris:04}
Andrew~Cameron Morris, Viktoria Maier, and Phil Green. 2004.
\newblock \href {https://doi.org/10.21437/Interspeech.2004-668} {{From WER and RIL to MER and WIL: improved evaluation measures for connected speech recognition}}.
\newblock In \emph{Interspeech 2004}, pages 2765--2768.

\bibitem[{Nariya et~al.(2023)Nariya, Mills, Sorger, and Sokolov}]{Nariya:23}
Maulik~K. Nariya, Caitlin~E. Mills, Peter~K. Sorger, and Artem Sokolov. 2023.
\newblock \href {https://doi.org/10.1016/j.patter.2023.100791} {{Paired evaluation of machine-learning models characterizes effects of confounders and outliers}}.
\newblock \emph{Patterns}, 4(8):100791.

\bibitem[{Negahban et~al.(2017)Negahban, Oh, and Shah}]{Negahban:17}
Sahand Negahban, Sewoong Oh, and Devavrat Shah. 2017.
\newblock \href {https://doi.org/10.1287/opre.2016.1534} {{Rank Centrality: Ranking from Pairwise Comparisons}}.
\newblock \emph{Operations Research}, 65(1):266--287.

\bibitem[{Newman(2023)}]{Newman:23}
Mark E.~J. Newman. 2023.
\newblock \href {http://jmlr.org/papers/v24/22-1086.html} {{Efficient Computation of Rankings from Pairwise Comparisons}}.
\newblock \emph{Journal of Machine Learning Research}, 24(238):1--25.

\bibitem[{Nguyen et~al.(2024)Nguyen, Seifert, and Schl{\"o}tterer}]{nguyen-etal-2024-ceval-benchmark}
Van~Bach Nguyen, Christin Seifert, and J{\"o}rg Schl{\"o}tterer. 2024.
\newblock \href {https://doi.org/10.18653/v1/2024.inlg-main.6} {{CE}val: A benchmark for evaluating counterfactual text generation}.
\newblock In \emph{Proceedings of the 17th International Natural Language Generation Conference}, pages 55--69, Tokyo, Japan. Association for Computational Linguistics.

\bibitem[{Pedregosa et~al.(2011)Pedregosa, Varoquaux, Gramfort, Michel, Thirion, Grisel, Blondel, Prettenhofer, Weiss, Dubourg, Vanderplas, Passos, Cournapeau, Brucher, Perrot, and Duchesnay}]{Pedregosa:11}
Fabian Pedregosa, Ga\"{e}l Varoquaux, Alexandre Gramfort, Vincent Michel, Bertrand Thirion, Olivier Grisel, Mathieu Blondel, Peter Prettenhofer, Ron Weiss, Vincent Dubourg, Jake Vanderplas, Alexandre Passos, David Cournapeau, Matthieu Brucher, Matthieu Perrot, and \'{E}douard Duchesnay. 2011.
\newblock \href {https://jmlr.org/papers/v12/pedregosa11a.html} {{Scikit-learn: Machine Learning in Python}}.
\newblock \emph{Journal of Machine Learning Research}, 12(85):2825--2830.

\bibitem[{Popovi{\'c}(2015)}]{popovic-2015-chrf}
Maja Popovi{\'c}. 2015.
\newblock \href {https://doi.org/10.18653/v1/W15-3049} {chr{F}: character n-gram {F}-score for automatic {MT} evaluation}.
\newblock In \emph{Proceedings of the Tenth Workshop on Statistical Machine Translation}, pages 392--395, Lisbon, Portugal. Association for Computational Linguistics.

\bibitem[{Post(2018)}]{post-2018-call}
Matt Post. 2018.
\newblock \href {https://doi.org/10.18653/v1/W18-6319} {A call for clarity in reporting {BLEU} scores}.
\newblock In \emph{Proceedings of the Third Conference on Machine Translation: Research Papers}, pages 186--191, Brussels, Belgium. Association for Computational Linguistics.

\bibitem[{Socher et~al.(2013)Socher, Perelygin, Wu, Chuang, Manning, Ng, and Potts}]{socher-etal-2013-recursive}
Richard Socher, Alex Perelygin, Jean Wu, Jason Chuang, Christopher~D. Manning, Andrew Ng, and Christopher Potts. 2013.
\newblock \href {https://aclanthology.org/D13-1170/} {Recursive deep models for semantic compositionality over a sentiment treebank}.
\newblock In \emph{Proceedings of the 2013 Conference on Empirical Methods in Natural Language Processing}, pages 1631--1642, Seattle, Washington, USA. Association for Computational Linguistics.

\bibitem[{Spearman(1904)}]{Spearman:04}
Charles Spearman. 1904.
\newblock \href {https://doi.org/10.2307/1412159} {{The Proof and Measurement of Association between Two Things}}.
\newblock \emph{The American Journal of Psychology}, 15(1):72--101.

\bibitem[{Srivastava et~al.(2023)Srivastava, Rastogi, Rao, Shoeb, Abid, Fisch, Brown, Santoro, Gupta, Garriga-Alonso, Kluska, Lewkowycz, Agarwal, Power, Ray, Warstadt, Kocurek, Safaya, Tazarv, Xiang, Parrish, Nie, Hussain, Askell, Dsouza, Slone, Rahane, Iyer, Andreassen, Madotto, Santilli, Stuhlm{\"u}ller, Dai, La, Lampinen, Zou, Jiang, Chen, Vuong, Gupta, Gottardi, Norelli, Venkatesh, Gholamidavoodi, Tabassum, Menezes, Kirubarajan, Mullokandov, Sabharwal, Herrick, Efrat, Erdem, Karaka{\c{s}}, Roberts, Loe, Zoph, Bojanowski, {\"O}zyurt, Hedayatnia, Neyshabur, Inden, Stein, Ekmekci, Lin, Howald, Orinion, Diao, Dour, Stinson, Argueta, Ferri, Singh, Rathkopf, Meng, Baral, Wu, Callison-Burch, Waites, Voigt, Manning, Potts, Ramirez, Rivera, Siro, Raffel, Ashcraft, Garbacea, Sileo, Garrette, Hendrycks, Kilman, Roth, Freeman, Khashabi, Levy, Gonz{\'a}lez, Perszyk, Hernandez, Chen, Ippolito, Gilboa, Dohan, Drakard, Jurgens, Datta, Ganguli, Emelin, Kleyko, Yuret, Chen, Tam, Hupkes, Padmakumar, Buzan, Mollo, Yang, Lee,
  Schrader, Shutova, Cubuk, Segal, Hagerman, Barnes, Donoway, Pavlick, Rodol{\`a}, Lam, Chu, Tang, Erdem, Chang, Chi, Dyer, Jerzak, Kim, Manyasi, Zheltonozhskii, Xia, Siar, Mart{\'\i}nez-Plumed, Happ{\'e}, Chollet, Rong, Mishra, Winata, de~Melo, Kruszewski, Parascandolo, Mariani, Wang, Jaimovitch-Lopez, Betz, Gur-Ari, Galijasevic, Kim, Rashkin, Hajishirzi, Mehta, Bogar, Shevlin, Schuetze, Yakura, Zhang, Wong, Ng, Noble, Jumelet, Geissinger, Kernion, Hilton, Lee, Fisac, Simon, Koppel, Zheng, Zou, Kocon, Thompson, Wingfield, Kaplan, Radom, Sohl-Dickstein, Phang, Wei, Yosinski, Novikova, Bosscher, Marsh, Kim, Taal, Engel, Alabi, Xu, Song, Tang, Waweru, Burden, Miller, Balis, Batchelder, Berant, Frohberg, Rozen, Hernandez-Orallo, Boudeman, Guerr, Jones, Tenenbaum, Rule, Chua, Kanclerz, Livescu, Krauth, Gopalakrishnan, Ignatyeva, Markert, Dhole, Gimpel, Omondi, Mathewson, Chiafullo, Shkaruta, Shridhar, McDonell, Richardson, Reynolds, Gao, Zhang, Dugan, Qin, Contreras-Ochando, Morency, Moschella, Lam, Noble,
  Schmidt, He, Oliveros-Col{\'o}n, Metz, Senel, Bosma, Sap, Ter~Hoeve, Farooqi, Faruqui, Mazeika, Baturan, Marelli, Maru, Ramirez-Quintana, Tolkiehn, Giulianelli, Lewis, Potthast, Leavitt, Hagen, Schubert, Baitemirova, Arnaud, McElrath, Yee, Cohen, Gu, Ivanitskiy, Starritt, Strube, Sw{\k{e}}drowski, Bevilacqua, Yasunaga, Kale, Cain, Xu, Suzgun, Walker, Tiwari, Bansal, Aminnaseri, Geva, Gheini, T, Peng, Chi, Lee, Krakover, Cameron, Roberts, Doiron, Martinez, Nangia, Deckers, Muennighoff, Keskar, Iyer, Constant, Fiedel, Wen, Zhang, Agha, Elbaghdadi, Levy, Evans, Casares, Doshi, Fung, Liang, Vicol, Alipoormolabashi, Liao, Liang, Chang, Eckersley, Htut, Hwang, Mi{\l}kowski, Patil, Pezeshkpour, Oli, Mei, Lyu, Chen, Banjade, Rudolph, Gabriel, Habacker, Risco, Milli{\`e}re, Garg, Barnes, Saurous, Arakawa, Raymaekers, Frank, Sikand, Novak, Sitelew, Le~Bras, Liu, Jacobs, Zhang, Salakhutdinov, Chi, Lee, Stovall, Teehan, Yang, Singh, Mohammad, Anand, Dillavou, Shleifer, Wiseman, Gruetter, Bowman, Schoenholz, Han,
  Kwatra, Rous, Ghazarian, Ghosh, Casey, Bischoff, Gehrmann, Schuster, Sadeghi, Hamdan, Zhou, Srivastava, Shi, Singh, Asaadi, Gu, Pachchigar, Toshniwal, Upadhyay, Debnath, Shakeri, Thormeyer, Melzi, Reddy, Makini, Lee, Torene, Hatwar, Dehaene, Divic, Ermon, Biderman, Lin, Prasad, Piantadosi, Shieber, Misherghi, Kiritchenko, Mishra, Linzen, Schuster, Li, Yu, Ali, Hashimoto, Wu, Desbordes, Rothschild, Phan, Wang, Nkinyili, Schick, Kornev, Tunduny, Gerstenberg, Chang, Neeraj, Khot, Shultz, Shaham, Misra, Demberg, Nyamai, Raunak, Ramasesh, prabhu, Srivastava, Padmakumar, Srikumar, Fedus, Saunders, Zhang, Vossen, Ren, Tong, Zhao, Wu, Shen, Yaghoobzadeh, Lakretz, Song, Bahri, Choi, Yang, Hao, Chen, Belinkov, Hou, Hou, Bai, Seid, Zhao, Wang, Wang, Wang, and Wu}]{Srivastava:23}
Aarohi Srivastava, Abhinav Rastogi, Abhishek Rao, Abu Awal~Md Shoeb, Abubakar Abid, Adam Fisch, Adam~R. Brown, Adam Santoro, Aditya Gupta, Adri{\`a} Garriga-Alonso, Agnieszka Kluska, Aitor Lewkowycz, Akshat Agarwal, Alethea Power, Alex Ray, Alex Warstadt, Alexander~W. Kocurek, Ali Safaya, Ali Tazarv, and 432 others. 2023.
\newblock \href {https://openreview.net/forum?id=uyTL5Bvosj} {{Beyond the Imitation Game: Quantifying and extrapolating the capabilities of language models}}.
\newblock \emph{Transactions on Machine Learning Research}, 5.

\bibitem[{Ustalov(2025)}]{ustalov-2025-reliable}
Dmitry Ustalov. 2025.
\newblock \href {https://aclanthology.org/2025.coling-demos.6/} {Reliable, reproducible, and really fast leaderboards with evalica}.
\newblock In \emph{Proceedings of the 31st International Conference on Computational Linguistics: System Demonstrations}, pages 46--53, Abu Dhabi, UAE. Association for Computational Linguistics.

\bibitem[{Wang et~al.(2019)Wang, Singh, Michael, Hill, Levy, and Bowman}]{Wang:19}
Alex Wang, Amanpreet Singh, Julian Michael, Felix Hill, Omer Levy, and Samuel~R. Bowman. 2019.
\newblock \href {https://openreview.net/forum?id=rJ4km2R5t7} {{GLUE: A Multi-Task Benchmark and Analysis Platform for Natural Language Understanding}}.
\newblock In \emph{Proceedings of the 7th International Conference on Learning Representations (ICLR) 2019}.

\bibitem[{Wolf et~al.(2020)Wolf, Debut, Sanh, Chaumond, Delangue, Moi, Cistac, Rault, Louf, Funtowicz, Davison, Shleifer, von Platen, Ma, Jernite, Plu, Xu, Le~Scao, Gugger, Drame, Lhoest, and Rush}]{wolf-etal-2020-transformers}
Thomas Wolf, Lysandre Debut, Victor Sanh, Julien Chaumond, Clement Delangue, Anthony Moi, Pierric Cistac, Tim Rault, Remi Louf, Morgan Funtowicz, Joe Davison, Sam Shleifer, Patrick von Platen, Clara Ma, Yacine Jernite, Julien Plu, Canwen Xu, Teven Le~Scao, Sylvain Gugger, and 3 others. 2020.
\newblock \href {https://doi.org/10.18653/v1/2020.emnlp-demos.6} {Transformers: State-of-the-art natural language processing}.
\newblock In \emph{Proceedings of the 2020 Conference on Empirical Methods in Natural Language Processing: System Demonstrations}, pages 38--45, Online. Association for Computational Linguistics.

\end{thebibliography}

\clearpage\appendix\onecolumn

\section{\label{app:jigsaw}Jigsaw Rankings}

We present below the scores of the described models from our Jigsaw-derived dataset \citep{Adams:17}.

\subsection{Raw Jigsaw Dataset (Section~\ref{sec:rq1})}

\begin{tabular}{lccccc}
\toprule
\textbf{Model} & \textbf{Acc} & \textbf{AUC} & \textbf{BT} & \textbf{F\textsubscript{1}} & \textbf{BT\textsubscript{bin}} \\
\midrule
TTA + PL & $0.895$ & $0.954$ & $0.082$ & $0.740$ & $0.122$ \\
JMTC-20 & $0.895$ & $0.955$ & $0.083$ & $0.739$ & $0.121$ \\
XLM-R & $0.889$ & $0.952$ & $0.093$ & $0.714$ & $0.115$ \\
XLM-RoBERTa & $0.886$ & $0.944$ & $0.067$ & $0.721$ & $0.118$ \\
XLM-R Conv1D & $0.883$ & $0.943$ & $0.167$ & $0.731$ & $0.117$ \\
XLM-RoBERTa Bayesian & $0.849$ & $0.501$ & $0.029$ & $0.171$ & $0.110$ \\
DistilBERT & $0.835$ & $0.882$ & $0.144$ & $0.523$ & $0.105$ \\
NB-SVM & $0.821$ & $0.866$ & $0.071$ & $0.367$ & $0.102$ \\
XGBoost & $0.754$ & $0.745$ & $0.264$ & $0.572$ & $0.089$ \\
\bottomrule
\end{tabular}

\subsection{Binarized Jigsaw Dataset (Section~\ref{sec:rq2})}

\begin{tabular}{lcccc}
\toprule
\textbf{Model} & \textbf{Accuracy} & \textbf{ROC AUC} & \textbf{BT} & \textbf{F\textsubscript{1}} \\
\midrule
XGBoost & $0.754$ & $0.745$ & $0.062$ & $0.572$ \\
XLM-RoBERTa Bayes & $0.797$ & $0.501$ & $0.008$ & $0.171$ \\
NB-SVM & $0.812$ & $0.866$ & $0.013$ & $0.367$ \\
XLM-RoBERT & $0.816$ & $0.944$ & $0.013$ & $0.721$ \\
DistilBERT & $0.819$ & $0.882$ & $0.021$ & $0.523$ \\
XLM-R Conv1D & $0.834$ & $0.943$ & $0.023$ & $0.731$ \\
TTA + PL & $0.846$ & $0.954$ & $0.015$ & $0.740$ \\
JMTC-20 & $0.849$ & $0.955$ & $0.015$ & $0.739$ \\
XLM-R & $0.856$ & $0.952$ & $0.017$ & $0.714$ \\
Binarized XGBoost & $0.754$ & $0.745$ & $0.060$ & $0.572$ \\
Binarized NB-SVM & $0.821$ & $0.612$ & $0.079$ & $0.367$ \\
Binarized DistilBERT & $0.835$ & $0.681$ & $0.081$ & $0.523$ \\
Binarized XLM-RoBERTa Bayes & $0.849$ & $0.499$ & $0.089$ & $0.171$ \\
Binarized XLM-R Conv1D & $0.883$ & $0.819$ & $0.100$ & $0.731$ \\
Binarized XLM-RoBERT & $0.886$ & $0.804$ & $0.099$ & $0.721$ \\
Binarized XLM-R & $0.889$ & $0.791$ & $0.099$ & $0.714$ \\
Binarized 1st Place & $0.895$ & $0.813$ & $0.104$ & $0.740$ \\
Binarized JMTC-20 & $0.895$ & $0.811$ & $0.101$ & $0.739$ \\
\bottomrule
\end{tabular}

\clearpage

\subsection{Penalized Jigsaw Dataset (Section~\ref{sec:rq2})}

\begin{tabular}{lcccc}
\toprule
\textbf{Model} & \textbf{Acc} & \textbf{AUC} & \textbf{BT} & \textbf{F\textsubscript{1}} \\
\midrule
XGBoost & $0.754$ & $0.745$ & $0.142$ & $0.572$ \\
XLM-RoBERTa Bayesian & $0.797$ & $0.501$ & $0.017$ & $0.171$ \\
NB-SVM & $0.812$ & $0.866$ & $0.040$ & $0.367$ \\
XLM-RoBERT & $0.816$ & $0.944$ & $0.032$ & $0.721$ \\
DistilBERT & $0.819$ & $0.882$ & $0.079$ & $0.523$ \\
XLM-R Conv1D & $0.834$ & $0.943$ & $0.088$ & $0.731$ \\
TTA + PL & $0.846$ & $0.954$ & $0.042$ & $0.740$ \\
JMTC-20 & $0.849$ & $0.955$ & $0.044$ & $0.739$ \\
XLM-R & $0.856$ & $0.952$ & $0.053$ & $0.714$ \\
Penalized XLM-RoBERTa Bayesian & $0.751$ & $0.502$ & $0.013$ & $0.171$ \\
Penalized XGBoost & $0.754$ & $0.745$ & $0.139$ & $0.572$ \\
Penalized XLM-RoBERT & $0.773$ & $0.625$ & $0.026$ & $0.721$ \\
Penalized DistilBERT & $0.787$ & $0.385$ & $0.065$ & $0.523$ \\
Penalized NB-SVM & $0.793$ & $0.228$ & $0.035$ & $0.367$ \\
Penalized XLM-R Conv1D & $0.793$ & $0.656$ & $0.072$ & $0.731$ \\
Penalized 1st Place & $0.812$ & $0.638$ & $0.034$ & $0.740$ \\
Penalized JMTC-20 & $0.816$ & $0.633$ & $0.036$ & $0.739$ \\
Penalized XLM-R & $0.827$ & $0.594$ & $0.045$ & $0.714$ \\
\bottomrule
\end{tabular}

\section{\label{app:sst5}SST-5 Rankings}

We present below the scores of the described models from the SST-5 dataset \citep{socher-etal-2013-recursive}.

\subsection{Raw SST-5 Dataset (Section~\ref{sec:rq1})}

\begin{tabular}{lcccc}
\toprule
\textbf{Model} & \textbf{Acc} & \textbf{BT} & \textbf{F\textsubscript{1}} \\
\midrule
TextBlob & $0.284$ & $0.067$ & $0.255$ \\
VADER & $0.316$ & $0.084$ & $0.315$ \\
Logistic Regression & $0.409$ & $0.135$ & $0.383$ \\
SVM & $0.414$ & $0.126$ & $0.401$ \\
\emph{fast}Text & $0.434$ & $0.120$ & $0.384$ \\
Flair-ELMo & $0.462$ & $0.143$ & $0.408$ \\
Transformer & $0.491$ & $0.162$ & $0.486$ \\
Flair-BERT & $0.511$ & $0.162$ & $0.491$ \\
\bottomrule
\end{tabular}

\subsection{Binarized SST-5 Dataset (Section~\ref{sec:rq1})}

\begin{tabular}{lcccc}
\toprule
\textbf{Model} & \textbf{Acc} & \textbf{BT} & \textbf{F\textsubscript{1}} \\
\midrule
TextBlob & $0.225$ & $0.032$ & $0.255$ \\
VADER & $0.248$ & $0.054$ & $0.315$ \\
Logistic Regression & $0.258$ & $0.043$ & $0.383$ \\
\emph{fast}Text & $0.272$ & $0.052$ & $0.384$ \\
Flair-ELMo & $0.344$ & $0.155$ & $0.408$ \\
Flair-BERT & $0.353$ & $0.124$ & $0.491$ \\
Transformer & $0.360$ & $0.154$ & $0.486$ \\
SVM & $0.384$ & $0.386$ & $0.401$ \\
\bottomrule
\end{tabular}

\clearpage

\section{\label{app:ceval}CEval Rankings}

We present below the scores of the described models from the CEval dataset \citep{nguyen-etal-2024-ceval-benchmark}.

\subsection{Raw CEval Dataset (Section~\ref{sec:rq2})}
\begin{tabular}{lcccc}
\toprule
\textbf{Model} & \textbf{ED} & \textbf{WER} & \textbf{chrF} & \textbf{BT}\\
\midrule
Crowd & $162.041$ & $0.239$ & $81.326$ & $0.444$ \\
MICE & $229.711$ & $0.299$ & $73.674$ & $0.163$ \\
Llama 2 & $274.370$ & $0.375$ & $70.886$ & $0.202$ \\
LLaMA & $298.368$ & $0.404$ & $68.378$ & $0.125$ \\
GDBA & $333.184$ & $0.540$ & $55.427$ & $0.017$ \\
Crest & $362.584$ & $0.477$ & $63.324$ & $0.049$ \\
\bottomrule
\end{tabular}

\subsection{Penalized CEval Dataset (Section~\ref{sec:rq2})}
\begin{tabular}{lcccc}
\toprule
\textbf{Model} & \textbf{ED} & \textbf{WER} & \textbf{chrF} & \textbf{BT}\\
\midrule
Crowd & $162.041$ & $0.239$ & $81.326$ & $0.240$ \\
MICE & $229.711$ & $0.299$ & $73.674$ & $0.093$ \\
Llama 2 & $274.370$ & $0.375$ & $70.886$ & $0.095$ \\
LLaMA & $298.368$ & $0.404$ & $68.378$ & $0.075$ \\
GDBA & $333.184$ & $0.540$ & $55.427$ & $0.025$ \\
Crest & $362.584$ & $0.477$ & $63.324$ & $0.023$ \\
Penalized Crowd & $272.713$ & $0.363$ & $79.950$ & $0.189$ \\
Penalized MICE & $384.359$ & $0.451$ & $72.188$ & $0.077$ \\
Penalized Llama 2 & $437.590$ & $0.592$ & $69.111$ & $0.078$ \\
Penalized LLaMA & $484.732$ & $0.657$ & $66.350$ & $0.059$ \\
Penalized GDBA & $475.117$ & $0.698$ & $54.434$ & $0.022$ \\
Penalized Crest & $458.033$ & $0.589$ & $62.539$ & $0.022$ \\
\bottomrule
\end{tabular}

\end{document}